%% file: RaveStatePaper.tex
\documentclass[
	a4paper, 
	10pt, 
]{LTJournalArticle}

\usepackage{caption}
\usepackage{mathtools} 
\usepackage{nicefrac}
\usepackage{epigraph}
\usepackage{etoolbox}
\usepackage{tcolorbox}
\usepackage{xspace}
\usepackage{thmtools}
\usepackage{framed}
\usepackage{hhline}
\usepackage{graphicx}
\usepackage{graphbox}
\usepackage{pgfplots}
\usepackage{tikz}
\usepackage{fancybox}
\usepackage[T1]{fontenc}
\usepackage{pythonhighlight}
\usepackage{hyperref}
\usepackage{multicol}
\usepackage{multirow}
\usepackage{tabularx}
\usepackage{makecell}
\usepackage{color, colortbl}
\hypersetup{
     colorlinks   = true,
     citecolor    = blue
}
\usepackage{enumitem}
\usepackage{amsmath}

\newtcbox{\textbox}[1][red]{on line,
arc=2pt,colback=#1!10!white,colframe=#1!50!black,
before upper={\rule[-2pt]{0pt}{9pt}},boxrule=0.5pt,
boxsep=0pt,left=2pt,right=2pt,top=1pt,bottom=.5pt}

\newcommand{\mychar}[1]{
  \begingroup\normalfont
  \includegraphics[height=1.5\fontcharht\font`\B]{#1}
  \endgroup
}

\newcommand{\rasta}{\textit{Ravestate}\xspace}
\newcommand{\roboy}{\textit{Roboy}\xspace}
\newcommand{\sigslot}{Signal-Rule-Slot}

\newcommand{\mathsc}[1]{\text{\textsc{#1}}}

\newcommand{\slot}[1][]{\ifstrempty{#1}{p}{\ensuremath{p_#1}}}
\newcommand{\Slots}{P}
\newcommand{\slotname}[1]{\textbox[cyan]{\textsc{#1}}}

\newcommand{\prodrule}[1][]{\ifstrempty{#1}{\ensuremath{\lambda}}{\ensuremath{\lambda_#1}}}
\newcommand{\Rules}{\ensuremath{\Lambda}}
\newcommand{\rulename}[1]{\textbox[red]{\textsc{#1}}}

\newcommand{\signal}[1][]{\ifstrempty{#1}{s}{\ensuremath{s_#1}}}
\newcommand{\Signals}{S}
\newcommand{\signame}[1]{\textbox[green]{\textsc{#1}}}

\newcommand{\Spikes}{\ensuremath{\Delta}}

\newcommand{\Activations}{A}

\newcommand{\module}[1]{\textbf{\pyth{#1}}}

\newtheorem{equ}{Equation}

\definecolor{wildtalk}{rgb}{0.78,0.93,1}
\definecolor{genqa}{rgb}{1,0.88,1}
\definecolor{persqa}{rgb}{1,1,0.88}
\definecolor{agentqa}{rgb}{0.86,0.82,1}
\definecolor{hibye}{rgb}{1, 0.6, 0.6}
\definecolor{emotion}{rgb}{0.56,0.93,0.56}

\setlist[itemize]{leftmargin=*,topsep=3pt,itemsep=-1ex,partopsep=1ex,parsep=1ex}
\setlist[enumerate]{leftmargin=*,topsep=3pt,itemsep=-3pt,partopsep=1ex,parsep=1ex}

\title{Ravestate: Distributed Composition of a Causal-Specificity-Guided Interaction Policy}

\author{
\begin{tabular}{ccc}
Birkner, Joseph & Dolp, Andreas & Karimi, Negin \\
\texttt{joseph.birkner@tum.de} & \texttt{andreas.dolp@tum.de} & \texttt{negin.karimi@tum.de}  \\[\bigskipamount]
Basargin, Nikita & Kharchenko, Alona & Hostettler, Rafael \\
\texttt{n.basargin@tum.de} & \texttt{alona.kharchenko@tum.de} & \texttt{rh@roboy.org} \\
\end{tabular}
}

\begin{document}

\maketitle 

\begin{abstract}
In human-robot interaction policy design, a rule-based method is efficient, explainable, expressive and intuitive. In this paper, we present the \sigslot~ framework, which refines prior work on rule-based symbol system design and introduces a new, Bayesian notion of interaction rule utility called \textit{Causal Pathway Self-information}. We offer a rigorous theoretical foundation as well as a rich open-source reference implementation \rasta, with which we conduct user studies in text-, speech-, and vision-based scenarios. The experiments show robust contextual behaviour of our probabilistically informed rule-based system, paving the way for more effective human-machine interaction.
\end{abstract}


\input{1_intro}

\input{2_prior}
\input{3_automaton}
\input{5_implementation}
\input{6_evaluation}
\input{7_conclusion}

\input{8_refs}
\appendix
\newpage
\end{document}

%% file: 1_intro.tex
\section{Introduction}

Non-physical, especially verbal and emotional interaction between humans and robots is envisioned to be fluent, context-sensitive and entertaining. To this end, a robot persona is designed with the agenda to encourage users to conduct certain specific activities with it, such as playing games, taking a selfie or buying ice cream. However, given the unbounded input space of natural language, it is hard to "insert" such an agenda into an interaction policy to facilitate a continuous flow of conversation. Ultimately, conversational disfluency events where the intent of a user's input is disrespected must be minimised while optimising for a series of diverse interactions that showcase specific distinct capabilities.\par
Understanding the intent of an interlocutor must take a mixture of the current and previous input signals into account, the result of which may again be a higher-level input signal which adds contextual information \cite{B94, rafaeli1988new}. Also, in a multi-modal system, different input sources (such as speech and vision) must be considered simultaneously.\par
One way to solve this problem is through end-to-end application of data-driven neural systems \cite{HKB+19}. However, these are very hard to "steer" from the perspective of a human interaction policy designer, and usually require them to express any necessary context for an action as a set of examples, which can neither be proven to be sufficient nor forced to be adequately generalised \cite{Goodfellow2015ExplainingAH, Kirkpatrick3521}.\par
Another approach are symbolic grammatical rule-based systems. For example, the rule $\{\text{\mychar{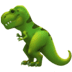}}\}\rightarrow\{\text{\mychar{figs/trex.png}\mychar{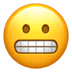}}\}$ describes how an agent develops anxiety at the sight of a Tyrannosaurus. Such rules exhibit a very large control surface to the interaction policy designer. For this reason, symbol production systems have been highly popular as mechanisms for the design of artificial agents in video games \cite{ruskin2012rule}. More importantly, they have been explored as architectures for cognitive modeling and inference \cite{W00}. \par

In this paper, we introduce \roboy's cognitive system \rasta\footnote{\url{https://github.com/roboy/ravestate}}, an open-source architecture for natural-language and vision-based interactive state machines which builds on prior work described in Section \ref{chap:prior}. Its architecture is based on the concept of a \sigslot~automaton, which is described in Section \ref{chap:design}. The \sigslot~automaton introduces a new probabilistic notion of action rule utility, which solves the problem of contextual action rule precedence and allows for dynamic combination of contextually overlapping actions. Interaction rules in this framework are productions which act as contextual modifiers of context, firing once a particular set of constraints in the current context is satisfied. This paper reveals the \sigslot~algorithm, the conflict resolution strategies between rules it employs (Section \ref{chap:exclusion}), its reference implementation \rasta (Section \ref{chap:impl}) and system evaluations through user studies (Section \ref{chap:eval}).\par
The reference implementation comes with a set of task- and modality-specific modular rule sets for verbal, visual and emotional interaction (see Figure \ref{fig:teaser}).

%% file: 2_prior.tex
\section{Prior Work}

\label{chap:prior}

The problem of mapping perception to action is fundamental to the field of artificial intelligence \cite{C03}. Through the Subsumption Architecture \cite{brooks1990elephants} it was discovered that by overlaying simple actions in response to some stimuli, complex life-like behavior of an artificial agent could be simulated.\par

\textit{SOAR} (State, Operator And Result) \cite{LHH+91} is a symbolic cognitive architecture where symbolic stimuli are modeled as so-called \textit{"working memory elements"}. Problem-solving occurs by matching rule left-hand sides against working memory elements in the current state. All matching rules are executed in parallel. The same applies to the TRINDI natural language dialogue toolkit \cite{larsson2000information}.\par

Another well-known symbolic cognitive architecture is \textit{ACT-R} (adaptive control of thought-rational) \cite{Anderson, ABB+04}. In ACT-R, rules (\textit{"processes"}) are executed in response to patterns of symbols (\textit{"chunks"}) which are placed in memory buffers. In contrast to SOAR, at each point in time, chunk patterns activate only a single rule, even if multiple left-hand rule patterns show a match. Only the rule  with the highest utility estimated via a Bayesian framework gets activated.\par

It should not go unmentioned that grammatical cognitive symbol systems also go by the name of "Blackboard architectures" \cite{van2006neural}, highlighting how a distributed rule-based cognitive architecture is based on independent rules operating over a shared working memory space (the blackboard). A truly multimodal blackboard-based interaction architecture was realised through the IrisTk \cite{skantze2012iristk} framework, which models state transitions in a hierarchical graph.\par

What we are missing from these existing architectures is a combination of the following properties:
\begin{enumerate}
    \item Distributed definition of rules.
    \item Parallel execution of matching rules.
    \item Intrinsic recognition of conflicting rules.
    \item Estimation of rule utility from causal structure.
\end{enumerate}

Perhaps closest to our needs, the OpenDial \cite{lison2016opendial} system allows for distributed specification of independently operating rules with utility derived from causal pathway probabilities through training, however no attempt is made to execute multiple matching rules in parallel or to derive intrinsic rule utility.\par

In the following we present a new architecture for rule-based automata covering the criteria above, which we implemented in the \rasta Python~3 library. 

%% file: 3_automaton.tex
\section{System Design}

\label{chap:design}

\subsection{Building Blocks}

The architecture of \rasta borrows concepts from both grammatical symbol systems and traditional Spoken-Language Understanding (SLU). SLU operates in terms of intents and slots \cite{tur2011spoken}. Slots are variables which are filled throughout a dialogue session to accomplish a goal which is guided by a detected intent. The combination of slot values and intent guides the action of the dialogue system. The following components make up \rasta's \sigslot~architecture:

\begin{enumerate}
    \item \textbf{Slots} are specific variables which accommodate the need to parameterise interaction state. For example, such variables could include the most recent user utterance and the system's verbal response in a dialogue application setting, or a wish expressed by the user, say strawberry ice cream, in a specific application.
    \item Symbols are adopted under the term \textbf{Signals}, highlighting their atomic and transient information storage capacity. Signals are events emitted deliberately through rules, or implicitly as "changed-events" when the value of a slot is changed. They are also outfitted with a payload value, which allows information transfer between a signal emitter and a consuming rule. Signals form the atoms of the left-hand rule condition parts.
    \item \textbf{Rules} are adopted as procedural memory elements which have the capacity to manipulate the signal pool and read/write slot values in response to a Boolean combination of signals, by executing a user-defined routine. In practice, a rule is simply an annotated function which is automatically executed by a rule-pattern matching loop.
\end{enumerate}

In the \rasta framework, a collection of symbols (working memory), slots (long-term memory) and rules (procedural memory) is called a \textit{context}. Note that \rasta does not yet support the formulation of arbitrary rule condition predicates over signal payload or slot values, although this would be a straight-forward addition.

\begin{figure}[h]
 \centering
 \includegraphics[width=\linewidth]{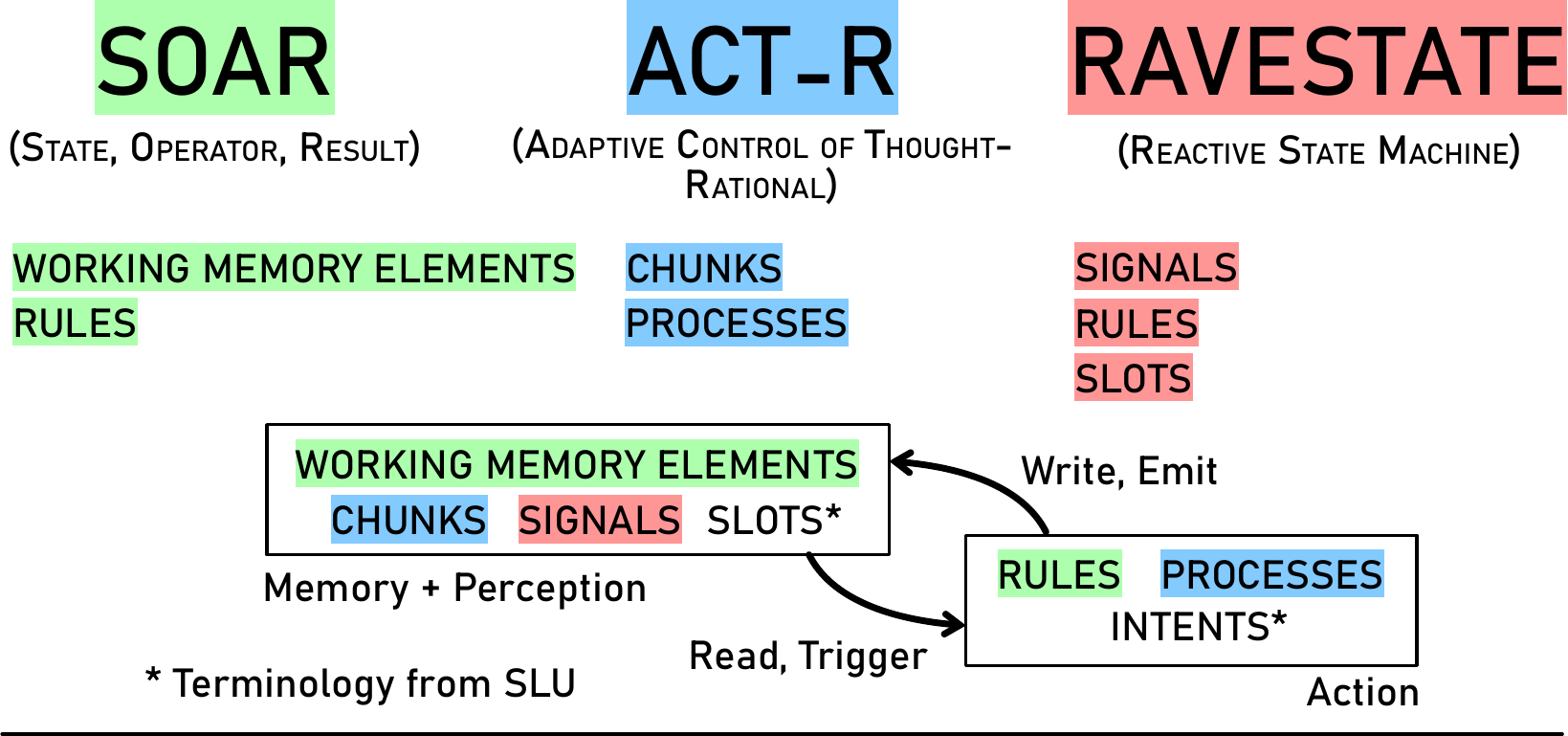}
 \caption{Terminology comparison regarding Perception-Action loops in \textit{SOAR}, \textit{ACT-R} and \rasta.}
 \label{fig:comparison}
\end{figure}

\subsection{\sigslot~Diagrams}

\label{appendix:diagrams}

\begin{figure}[h]
\centering
\begin{tabular}{|p{5em}|p{13em}|}
\hline
\includegraphics[scale=0.6,align=c,margin=0.1em]{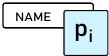} & \textbf{slot} \slot[i]\\\hline
\includegraphics[scale=0.6,align=c,margin=0.1em]{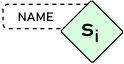} & \textbf{signal} \signal[i]\\\hline
\includegraphics[scale=0.6,align=c,margin=0.1em]{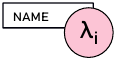} & \textbf{rule} \prodrule[i]\\\hline
\includegraphics[scale=0.6,align=c,margin=0.1em]{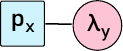} & \prodrule[y] \textbf{reads} \slot[x]\\\hline
\includegraphics[scale=0.6,align=c,margin=0.1em]{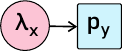} & \prodrule[x] \textbf{writes} to \slot[y]\\\hline
\includegraphics[scale=0.6,align=c,margin=0.1em]{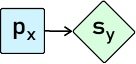} & \slot[x] \textbf{emits} \signal[y]\\\hline
\includegraphics[scale=0.6,align=c,margin=0.1em]{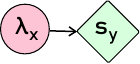} & \prodrule[x] \textbf{emits} \signal[y]\\\hline
\includegraphics[scale=0.6,align=c,margin=0.1em]{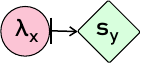} & \prodrule[x] \textbf{emits} \textbf{detached} \signal[y]\\\hline
\includegraphics[scale=0.6,align=c,margin=0.1em]{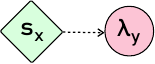} & \signal[x] is \textbf{condition} of \prodrule[y] \\\hline
\includegraphics[scale=0.6,align=c,margin=0.1em]{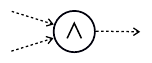} & \textbf{conjunction} of conditions\\\hline
\includegraphics[scale=0.6,align=c,margin=0.1em]{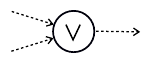} & \textbf{disjunction} of conditions\\\hline
\end{tabular}
\caption{\sigslot~Diagrams}
\label{fig:pss-diagrams}
\end{figure}

\sigslot~diagrams allow visualising relationships between slots, signals and rules (see Fig.~\ref{fig:pss-diagrams}).

\subsection{Parallel Rule Execution}

Much like in \textit{SOAR}, rules in \rasta are executed in parallel if they do not interfere with each other. The potential for colliding rules is explained in more detail in Section \ref{chap:exclusion}. Running rules in parallel has a number of practical advantages:

\begin{enumerate}
    \item \textbf{Long-running rules:} 
    Other rules are not blocked by a long-running rule function executing background tasks.
    \item \textbf{Fillers/Disfluencies:} In cases where response generation takes a few seconds, a filler rule voicing an utterance such as \textit{"Uhm..."} or \textit{"Let me think..."} is able to fill in the gap.
    \item \textbf{Heterogeneous output:} 
    In a multi-modal interactive system, some rules might complement each other positively when executed simultaneously, for example, by changing the facial expression and replying with a spoken remark, even without knowledge of mutual action.
    \item \textbf{High event frequency:} If specific signals are generated at a high frequency, it might be necessary to execute a rule again before the prior execution instance finishes.
\end{enumerate}

\subsection{Example}

\begin{figure}[h]
\centering
\vspace{3pt}
\includegraphics[width=0.5\linewidth]{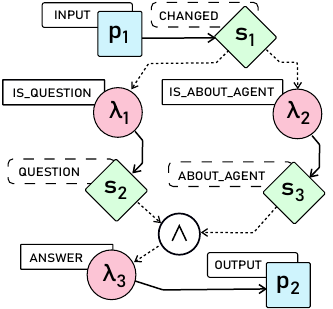}
\caption{\sigslot~diagram for simple conditioned question answering.}
\vspace{-10pt}
\label{fig:gra1}
\end{figure}

In combination, \textbf{sets of rules \Rules, signals \Signals, and slots $\Slots$} describe an agents response to an input. We provide an example for this is in Figure ~\ref{fig:gra1}: The slot \slotname{Input} contains a textual representation of a user utterance (e.g. \textit{"How are you?"}). As the value is placed into the slot, the \signame{Input:Changed}-Signal is triggered, which satisfies the condition for the \rulename{Is\_Question} and \rulename{Is\_About\_Agent} rules, which fire their respective \signame{Question} and \signame{About\_Agent} signals depending on the value of \slotname{Input}. The conjunction of these two signals triggers the personal question \rulename{Answer} rule, placing respective responses in the \slotname{Output} slot.

\textit{Note:} The exact elements of \sigslot-graphs as used in Figure \ref{fig:gra1} are described in Appendix \ref{appendix:diagrams}.

\subsection{Rule exclusion}

\label{chap:exclusion}

\begin{figure}[h]
\centering
\includegraphics[width=0.7\linewidth]{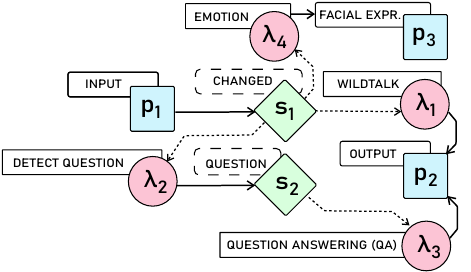}
\caption{Example for exclusive and non-exclusive rules: The rule $\prodrule[3]$ is racing with $\prodrule[1]$ to reply to signal $\signal[1]$, while $\prodrule[4]$ is independent. }
\vspace{-10pt}
\label{fig:gra2}
\end{figure}

In the previous example, all rules coexist and cooperate meaningfully simply by running as soon as their condition signals are available. However, it is easy to come up with a scenario where two rules would not cooperate in this way. Specifically, there may be two rules that overlap in their purpose to generate a behavior in response to a given stimulus, resulting in a bad combination of two response variants. The toy \sigslot~automaton in Figure \ref{fig:gra2} showcases such a situation. In this example there are two rules which may write to the \slotname{Output} slot; the \rulename{Wildtalk} rule and the \rulename{QuestionAnswering} (\rulename{Qa}) rule. The desired behavior is such that the \rulename{Qa} rule should apply when the input is a question, while \rulename{Wildtalk} (perhaps encapsulating a generative model) applies otherwise. As such, the two rules are designed to be mutually exclusive for each \signame{Input:Changed} signal, since executing both would entail a race condition. On the other hand, the \rulename{Emotion} rule which writes to \slotname{FacialExpr} can always run idependently from \rulename{Qa} or \rulename{Wildtalk}.\par

The example therefore illustrates how slots are devices which reduce the number of possible behaviors of a \sigslot~automaton. Specifically, all rules $\prodrule[x],\prodrule[y] \in \Rules$ like\par
\begin{center}
\signame{$\Signals_x$}~$\rightarrow$~
\rulename{$\prodrule[x]$}~$\rightarrow$~
\slotname{$\Slots_x$}
\end{center}
which write to a set of slots \slotname{$\Slots_x$} in reaction to a conjunction of signals \signame{$\Signals_x$} are mutually exclusive, if they share both written slots and condition signals ($(\Slots_x \cap \Slots_y \neq \emptyset) \land (\Signals_x \cap \Signals_y \neq \emptyset)$).\par
Note that this also applies to the \rulename{Wildtalk} and \rulename{Qa} rules in Figure \ref{fig:gra2}. Even though their immediate condition signal-sets \{~\signame{Input:Changed}~\} and \{~\signame{Question}~\} are exclusive, the \signame{Question} signal can actually only occur as an effect of \signame{Input:Changed}. Therefore, the complete condition of \rulename{Qa} is  \{~\signame{Input:Changed} $\land$ \signame{Question}~\}.

\subsection{Causal Pathway Utility}

\label{chap:utility}

Given the example above, one might simply solve the rule precedence problem by counting the condition signals of a rule, and prioritising a rule with more conditions over a rule with less. This metric is called \textit{specificity}, and it is a traditional component of rule-based systems \cite{mcdermott1978production}: The higher the specificity, the higher the situational utility and therefore the priority of a rule. Unfortunately, the basic counting metric easily breaks down in certain scenarios.\par

\begin{figure}[h]
\centering
\includegraphics[width=0.7\linewidth]{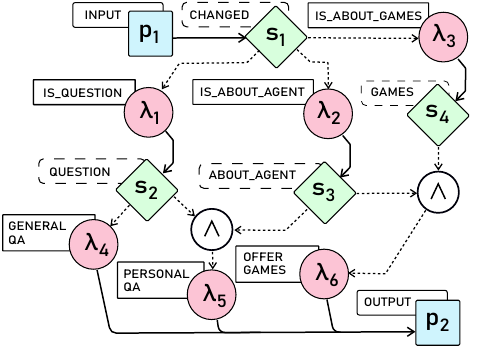}
\caption{Breakdown scenario of count-based specificity: \prodrule[5] and \prodrule[6] each have three condition signals.}
\vspace{-10pt}
\label{fig:gra3}
\end{figure}

The basic flaw of count-based specificity becomes apparent in the example presented by Figure \ref{fig:gra3}: The \rulename{OfferGames} and \rulename{PersonalQa} rules each have three signals in their condition conjunctions. Therefore, an input like \textit{"Do you like games?"} can not be decisively routed to either rule under count-based specificity. Revisiting the intuition of $\text{\textsc{Specificity}}\propto\text{\textsc{Utility}}$, the proportional relationship may be explained by the lower cross-entropy between a given situation and a rule which is modeled \textit{specifically} for this situation vs. a rule that is modeled to cover a broader spectrum of situations. Therefore, what we might actually want to model is simply the prior probability of a rule condition. This probability's negative logarithm is its self-information which correlates with specificity, giving us Equation \ref{eq:selfinfo}.

\begin{equ}
\begin{leftbar}
Given the self-information $H$ of a random event $x$ with probability $p(x)$ as $H(x)=-log(p(x))$, for any rule $\prodrule[i]$
$$\text{\textsc{Utility}}(\prodrule[i])\propto \text{\textsc{Specificity}}(\prodrule[i])\propto -\text{log}(p(\prodrule[i]))$$
\vspace{-10pt}
\label{eq:selfinfo}
\end{leftbar}
\end{equ}

We therefore need to find a way to calculate the probability estimate $p(\prodrule[i])$ for a rule $\prodrule[i]$, without any data from which a Maximum Likelihood Estimate may be derived. Using a single assumption, we can derive a probability estimate just from the structure of the automaton itself: \textit{The amount of the information carried by a signal is inversely proportional to the number of rules depending on it.} This assumption implies that

\begin{enumerate}
    \item A signal which occurs in every rule condition does not carry any information.
    \item A signal which occurs in only a single rule condition carries the highest possible amount of information.
\end{enumerate}

Since a logical conjunction corresponds to a product of probabilities which corresponds to a sum of self-information values, we can calculate the self-information of a rule as the sum of the self-information values of its conjunct condition signals. This is expressed in Equation \ref{eq:pathwayinfo}.

\begin{equ}
\begin{leftbar}
Given a \sigslot~automaton $(\Signals, \Rules, \Slots)$ consisting of slots $\Slots$, signals $\Signals$ and rules $\Rules$, the self-information of any single rule $\prodrule[i] \in \Rules$ with its conjunct condition clause $\Signals_i \subseteq \Signals$ is calculated as \\[\bigskipamount]
\begin{math}
\begin{array}{c@{\,}l}
\mathsc{Utility}(\prodrule[i]) & = \sum_{s \in \Signals_i} -\log\left(\frac{|\{s \in \Signals_k | \prodrule[k] \in \Rules \}|}{|\Rules|}\right) \\
\scriptstyle{= H(\prodrule[i])} & 
\end{array}
\end{math}
\label{eq:pathwayinfo}
\end{leftbar}
\end{equ}

Note, that this definition of utility introduces an optimization requirement on the interaction policy designer (a human for now) to model rule conditions such that there is minimal mutual information between a rule's condition's signals.

\subsection{Loops And Causal Groups}

Figure \ref{fig:gra2} in Section \ref{chap:exclusion} shows that the \textit{causal history} of a signal must contribute to its \textit{exclusion calculation} with other signals. From the example, the \signame{Question} signal is overlapping with \signame{Input:Changed} because it is \textit{caused} by \signame{Input:Changed} through the \rulename{Question} rule. We denote signals which overlap in their sets of recursive causes as belonging to the same \textit{causal group}. The clause $ \Signals_x \cap \Signals_y = \emptyset$ from Section \ref{chap:exclusion} therefore expresses that in order to be non-conflicting, two rules must have conditions which belong to separate causal groups.\par
This requirement, together with the requirement that any implementation of the \sigslot~system automatically performs cause-based recursive completion of condition signals, would greatly restrict the number of possible behaviors that can be expressed with a \sigslot~automaton: The impossible behaviors would be all those which causally originate from a previous behavior of the agent herself. Since this is a particularly interesting set of behaviors, the concept of \textit{detached emission} is introduced.

\begin{figure}[h]
\centering
\includegraphics[width=0.7\linewidth]{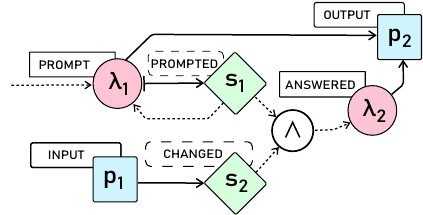}
\caption{Example for a "detached" signal emission scenario.}
\vspace{-10pt}
\label{fig:gra4}
\end{figure}

An example for detached emission is given in Figure \ref{fig:gra4}: Through some previous cause, the agent prompts the user with a question. This state is saved by the \signame{Prompted} signal. Through detached emission, \signame{Prompted} is not completed with the condition signals of the \rulename{Prompt} rule. Thereby, \signame{Prompted} can both trigger a loop of asking the question again after a certain time-out, or trigger the \rulename{Answered} rule in conjunction with \signame{Input:Changed}.

\subsection{Integration of Visual Stimuli}

\label{chap:visionio}

As far as specific mechanics are concerned, examples in Section \ref{chap:design} only showcase how \sigslot~automata (and by extension our \rasta reference implementation) function in scenarios based on verbal interaction. As a proof-of-concept, this Section contains an example of how vision-based interlocutor tracking and recognition (using face recognition in particular) is integrated into \rasta using the \sigslot~system.

\begin{figure}[h]
\centering
\includegraphics[width=\linewidth]{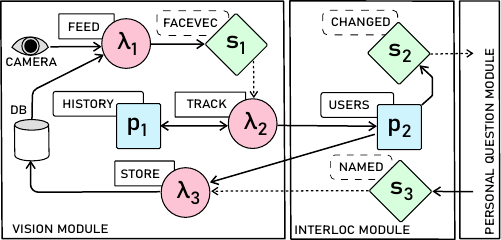}
\caption{The VisionIO module architecture.}
\vspace{-10pt}
\label{fig:gra5}
\end{figure}

Figure \ref{fig:gra5} shows how the \module{visionio} module interacts with the \module{interloc} and \module{persqa} modules to store and recall the present interlocutor based on her facial feature vector: The \rulename{Feed} rule continuously emits \signame{facevec} signals for facial feature vectors which are extracted from a generic camera input stream, and associated with primary keys from persistent graph memory if a close feature match based on cosine similarity is found. The \signame{facevec} feature signals are received by the \rulename{Track} rule, which tracks a history of such signals in a ring-buffer placed in the \slotname{history} slot. The ringbuffer debounces the \signame{facevec} signal, and allows the \rulename{track} rule to decide when confidence is high enough to update the \module{interloc} module's central \slotname{users} slot with a new graph database node that represents the new interlocutor. If the new user is unknown, their database node will be transient. Only once the name of the user is inferred by the \module{persqa} module, the node will be persisted, which is expressed in the \signame{named} signal. This allows the \rulename{store} rule to update the facial feature vector database.

\subsection{Spikes and Activations}

\begin{figure}
\centering
\includegraphics[width=0.9\linewidth]{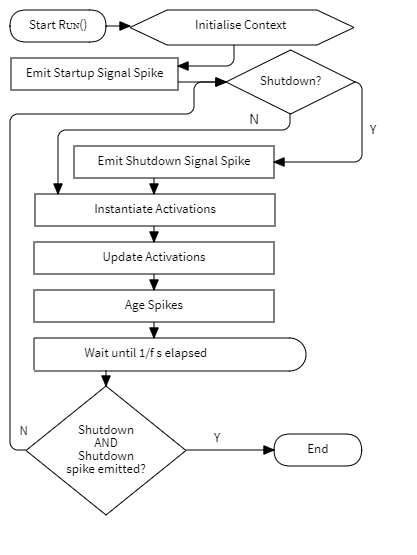}\vspace{-2em}
\caption{Context \textsc{Run} routine. The routine references a fixed update frequency $f$.}
\vspace{-1em}
\label{fig:context-run}
\end{figure}

\begin{figure}
\centering
\includegraphics[width=0.9\linewidth]{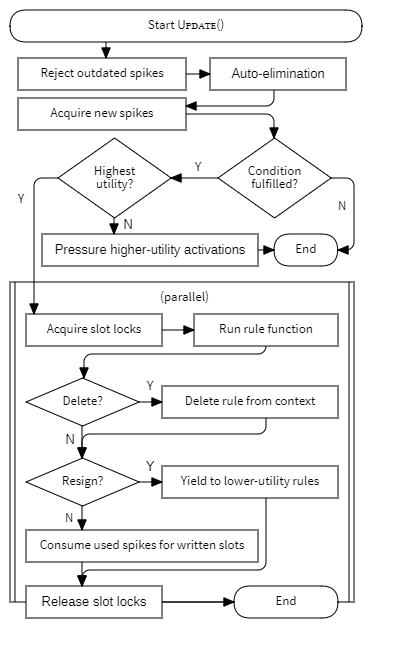}\vspace{-1.5em}
\caption{Activation \textsc{Update} routine. It is repeatedly called by a context's \textsc{Run} routine for each activation.}
\label{fig:act-update}
\end{figure}

We have described the declaration of \sigslot~automata, and how they can be used to model behavioral policies for interactive agents. We have however not specified the procedures by which the automaton's state is transitioned at runtime: Most importantly, we have not introduced \textit{spikes} and \textit{activations}, and the role they play in the system: \textit{Spikes}, consisting of a signal reference and an age, encapsulate instances of signals that are \textit{emitted} during the run of a rule function. \textit{Activations}, consisting of a rule reference and a set of acquired spikes, encapsulate all state that is associated with the execution of a rule.\par
A combination of rules \Rules, activations \Activations, signals \Signals, spikes $\Spikes$ and slots $\Slots$ is aggregated into a \textit{context} structure $C = \langle\Rules,\Signals,\Slots,\Spikes,\Activations\rangle$. Such a context fully describes a single \sigslot~system at any given point in time. Its temporal transition follows the \textsc{Run} routine (see Fig.~\ref{fig:context-run}).\par
The \textsc{Run} routine creates activations during the step \textit{"Instantiate Activations"}. For its activation collection $\Activations$, the context guarantees that for every signal $\signal[i]$ in a condition of one of its rules $\prodrule[i]$, there exists a "standby" activation which does \textit{not} hold a spike which fulfills said signal.\par
Each activation that is held in a context is continuously updated using the \textsc{Update} function (see Fig.~\ref{fig:act-update}) to facilitate the Activation's lifecycle: Activations live until they have acquired a sufficient spike set fulfilling their rule’s condition. Once an activation has acquired the necessary spikes, it may execute its rule function and remove itself from its context's activation set.\par
The \textsc{Update} algorithm also includes a mechanic which ensures that only the activation that promises the highest utility for its combination of held signal spikes and write-slots is executed: Once ready to run, the activation determines whether to go forward through a highest-utility check. If the activation's rule does not promise the highest available utility, unfulfilled higher-utility activations are \textit{pressured}.\par
A \textit{pressured activation} is subject to a so-called \textit{auto-elimination} process: It must either gather sufficient signal spikes to run within a certain timeout window, or reject its previously acquired spikes. This way, either a high-utility rule activation gains precedence over a lower-utility one, or it eventually yields when the specific signals spikes it requires are not becoming available.\par
Yielding to lower-utility activations may also be achieved explicitly by returning \textsc{Resign} from a rule function, indicating that the rule could not be applied for any reason. If the executed activation does not resign, it forces competing lower-utility activations to give up on the used spikes by \textit{consuming} them for the written slots.\par
The \textsc{Update} routine also enables time-based rule signal constraints: Signal constraints for rules may be annotated with a minimum and a maximum age for matching spikes. While an acquired spike is too young, the condition-fulfilled check will fail. If the spike becomes too old, it is rejected at the beginning of the \textsc{Update} routine.

%% file: 5_implementation.tex
\begin{figure*}[h]
    \includegraphics[width=\textwidth]{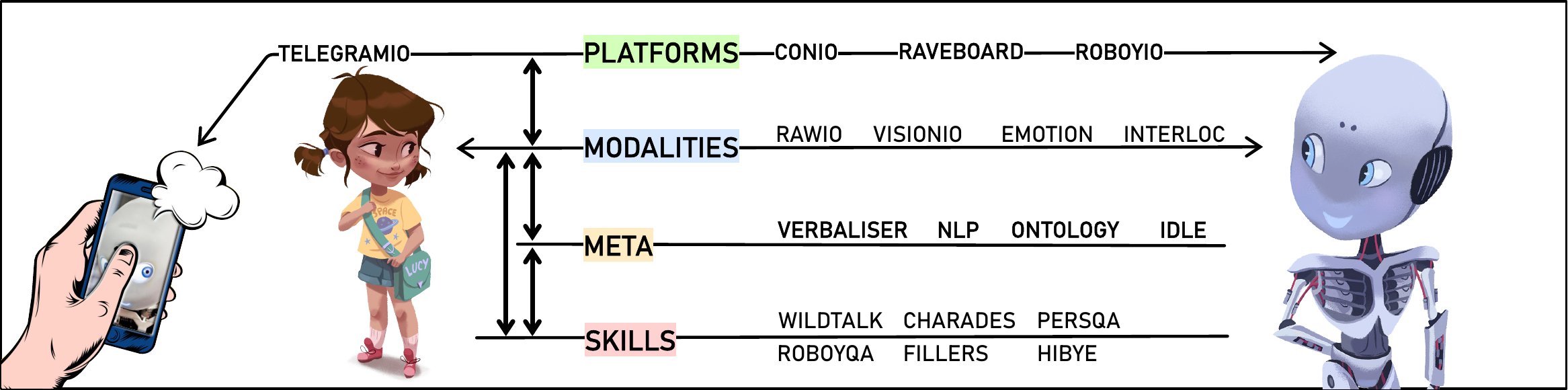}
    \caption{Our open-source reference implementation of a causal-specificity-aware automaton comes with modular sets of interaction rules which allow composing a behavior policy from independent building blocks. These building blocks are divided into categories based on their role in the system. The categories are \textbf{Platform Modules}, \textbf{Modality Modules}, \textbf{Meta Modules} and \textbf{Skill Modules}.}
    \label{fig:teaser}
\end{figure*}

\newpage
\section{Implementation}
\label{chap:impl}

We open-source our reference implementation of the \sigslot~automaton in the \rasta Python~3 library, together with modular sets of rules, slots and signals which allow composing an interactive agent from separate independent building blocks. These building blocks are divided into categories based on their role in the system, as shown in Figure \ref{fig:teaser}. The categories are \textbf{Platform Modules}, \textbf{Modality Modules}, \textbf{Meta Modules} and \textbf{Skill Modules}. Each module is generally structured as a separate Python package.

\subsection{Platform Modules}

Platform modules provide bindings to specific platforms for certain modalities. Implementations of speech, emotion expression or vision input all require specialised middleware for specific platforms like instant messaging, the physical robot and debug tools:

\begin{itemize}
    \item The web platform \module{webio} allows interaction with a \rasta agent through a chat interface, while inspecting signal spikes and activations as they occur in real-time. A more basic implementation for text-based interaction is offered through the \module{conio} module, which implements chatting on the command line.
    \item For instant-messaging, the \module{telegramio} module supports textual interaction with \rasta agents on the Telegram platform.
    \item The \module{roboyio} module provides platform bindings to physical instances of the \roboy robot. The robotic platform supports conversational interaction through speech-recognition and synthesis, facial expressions, head movement and face recognition.
    \item Adding more platforms is simplified through the \module{ros1} and \module{ros2}\footnote{\url{https://www.ros.org/}} modules, which provide specialised slot types that can bind to publish-subscribe network communication topics.
\end{itemize}

Importantly, the nature of the \sigslot~architecture allows for these modules to run in parallel without interfering with each other. For example, \module{webio} and \module{roboyio} can naturally co-exist at runtime.

\begin{figure}[h]
 \centering
 \includegraphics[width=1\linewidth]{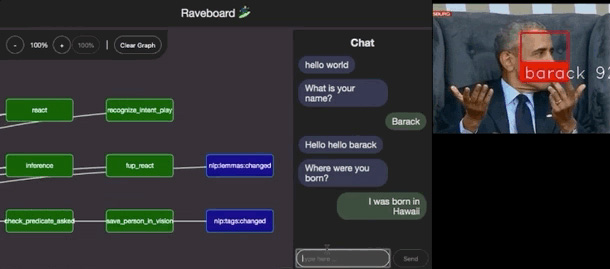}
 \caption{Raveboard UI with \module{visionio}}
 \label{fig:raveboard}
 \vspace{-10pt}
\end{figure}

\subsection{Modality Modules}

Modality modules provide platform-independent I/O interfaces for specific modalities to upstream platforms and downstream meta processors or skills:

\begin{itemize}
    \item The \module{rawio} module provides the \slotname{rawio:in} and \slotname{rawio:out} slots for bidirectional raw textual utterances. \par
    \item For visual interaction, the \module{visionio} module provides face recognition and interlocutor tracking based on a generic stream of input images (see Section \ref{chap:visionio}).
    \item Emotional I/O is provided through the \module{emotion} module, which offers emotion-specific signals that can be used by downstream applications.
    \item Perhaps an odd-one-out, the \module{interloc} module provides a standardised way for platforms and applications of accessing present interlocutors through the \slotname{users} slot.
\end{itemize}

\subsection{Meta Modules}

Meta modules provide middleware components that execute generic application-independent I/O refinements or expose auxiliary tools to downstream applications:

\begin{itemize}
    \item The \module{verbaliser} module offers an abstract way for applications to produce verbal responses, by writing an intent key to the \slotname{verbaliser:intent} slot which is then translated into a phrase.
    \item For advanced textual input processing the \module{nlp} module provides various signals and slots that are updated with output from the \textit{spacy} library \cite{spacy2} for subject-predicate-object triple extraction, sentence type detection, named entity recognition and other tasks.
    \item The \module{idle} module provides the \signame{idle:bored} and \signame{idle:impatient} signals that simplify the detection of "awkward pauses" in the interaction, due to processing lag or lack of user engagement.
    \item Long-term persistent memory is provided by the \module{ontology} module through a slot type that automatically binds values to nodes in a graph database.
\end{itemize}

\subsection{Skill Modules}

Skill modules provide implementations for specific interactive capabilities that can be cherry-picked to assemble the behavioral components of an artificial agent:

\begin{itemize}
    \item Neural generative response models are provided through the \module{wildtalk} module with \textit{ConvAI} and \textit{GPT-2} \cite{convAI}, and the \module{genqa} module for general knowledge question answering based on DrQA \cite{chen2017reading}.
    \item The \module{charades} module implements the charades game which involves guessing a user activity from visual cues, voicing the guess and reacting to the user's judgement.
    \item Greeting and farewell triggered by changes in the \slotname{interloc:users} slot are covered by the \module{hibye} module.
    \item The \module{persqa} module implements personalized smalltalk to store and recall trivia like name, hobby or occupation of an interlocutor.
    \item Through \module{agentqa}, an application can provide the interlocutors with answers to basic personal questions about the dialogue agent.
    \item The \module{fillers} module activates on \signame{idle:impatient} to utter conversational filler words when processing an utterance is taking longer.
\end{itemize}

%% file: 6_evaluation.tex
\section{User trials}

\label{chap:eval}

We conducted two user studies in the scope of Bachelor's theses, focusing on various aspects of human-machine interaction policy design, using the robot \roboy \cite{Trendel}. Both were implemented using the \rasta framework. Experimental results aided in the evaluation of the proposed system's features, such as modularity and the ability to facilitate multi-modal communication.

\subsection{Study A}

In the first study (\textit{Study A}), 20 users were offered to conduct an \textit{open-domain dialog} with a chatbot in textual form and with a humanoid robot as a spoken conversation. In the first part of the experiment only \rasta's language-related input and output modalities were enabled: text messages for the chatbot and speech recognition and synthesis for physical robot. In the second part of the experiment, with the goal to assess the participants' perception of the dialogue with additional modalities, we extended \rasta's outputs to include facial expressions (e.g. surprised, shy, affectionate) and head movements for the physical robot, and emojis for the chatbot. Every person spent 5 minutes conversing with the chatbot and 5 minutes talking to \roboy in person. Note that the utterance-response examples from Appendix \ref{appendix:flows} originate from this study.

\input{A0_dolp}

\subsection{Study B}

The second study (\textit{Study B}) was designed as a \textit{goal-oriented interaction} between the physical robot and a user. Here, we observed 18 users playing charades - a word guessing game - with the physical robot. We evaluated user responses to a new modality: visual feedback. A real-time human activity recognition system was integrated with \rasta which provides information about the current action of the interlocutor. In addition to visual feedback, speech recognition, speech synthesis and facial expression I/O modules were enabled. On average participants spent 8 minutes 36 seconds playing with the robot. Since the interaction with \rasta in this experiment was exclusively embodied, we employed metrics from human-robot interaction research and evaluated the agent along the dimensions of perceived intelligence, likability, sociability and animacy among other characteristics.

\input{A1_fedoseeva}

\subsection{Sample Conversations}

\input{A4_sample_conv}

\subsection{Discussion of Results}

Both studies demonstrated the ability of the proposed framework to cope with a variety of modalities: Written language and spoken language, as well as bidirectional visual feedback in the form of facial expressions display and vision-based human activity recognition. In \textit{Study A}, participants ranked various aspects of the interaction on a Likert scale from 1 (not at all accurate) to 5 (extremely accurate). The conversation with the chatbot ($\mu=2.10$, $\sigma=0.88$) was deemed to be more similar to another human than the conversation with the robot ($\mu=1.50$, $\sigma=0.97$). Similarly, the rankings for \textit{"The system understood what I said"} statement are lower for the robot ($\mu=2.10$, $\sigma=0.74$) vs. chatbot ($\mu=2.90$, $\sigma=0.74$) - which can be partially attributed to unreliable automatic speech recognition. \textit{Study A} respondents found \roboy rather friendly in both parts of the experiment ($\mu=3.98$, $\sigma=0.8$). According to the survey results, \textit{Study B} participants clearly noticed the newly added modalities and, based on our observations, spent more time looking at the robot's face. Moreover, \textit{Study B} showed that integrating visual feedback into the agent's responses increases perceived friendliness even more ($\mu=4.5$, $\sigma=0.71$). Similar to \textit{Study A}, in the second experiment all of participants also noticed the changes in \roboy's facial expressions and were able to provide explanations for them. Furthermore, 94\% of the respondents reported that they experienced empathy (35.29\%), joy (29.41\%) or slight frustration (23.53\%) while playing charades with the robot. Both studies were conducted with \rasta, indicating the framework's utility as a design tool for in-depth HRI research experiments.

%% file: A0_dolp.tex

Participants answered 14 questions on a five-point Likert scale from "1 - Not at all accurate" to "5 - Extremely accurate" for both conversations. The keywords "robot" and "chatbot" have been replaced by \texttt{system}. The resulting means and standard deviations of the answers to the five-point scale questions are shown in Table~\ref{fig:study_results}.

\begin{table}[!h]
\centering
\begin{tabular}{|p{0.65\linewidth}|c|c|}
\hline
Question & $\mu$ All & $\sigma$ All\\\hline\hline
The \texttt{system} understood what I said. &
 2.65 & 0.80 \\\hline
I understood what the \texttt{system} said.\footnote{Only relevant for speech modality.} &
 3.15 & 0.67 \\\hline
I was anxious during the conversation. &
 1.57 & 0.87 \\\hline
The \texttt{system}'s answers were out of context. &
 3.25 & 0.95 \\\hline
The utterances of the \texttt{system} were unvaried . &
 2.55 & 0.88 \\\hline
The \texttt{system} was repeating itself. &
 2.77 & 1.23 \\\hline
The \texttt{system} showed some kind of emotion. &
 2.90 & 1.15 \\\hline
The \texttt{system} was friendly. &
 3.98 & 0.80 \\\hline
The conversation felt slow. &
 3.27 & 1.43 \\\hline
The conversation was tedious. &
 2.90 & 0.96 \\\hline
I was able to gather new information in the conversation. &
 2.58 & 1.28 \\\hline
It felt similar to a dialog with another human. &
 1.85 & 0.92 \\\hline
I enjoyed the conversation. &
 3.60 & 1.01 \\\hline
My overall impression of the conversation was good. &
 3.15 & 0.92\\
\hline
\end{tabular}
\caption{\textit{Study A} survey results}
\label{fig:study_results}
\end{table}

%% file: A1_fedoseeva.tex

The \textit{Study B} survey uses a semantic differential scale, which provides a scale from 1 to 5 with so-called anchors on each end (words with the opposite meaning). The questionnaire measures five key concepts of HRI: anthropomorphism (\textbf{ATPR}), animacy (\textbf{ANIM}), likability (\textbf{LIKB}), perceived intelligence (\textbf{PINT}), and perceived safety (\textbf{PSAF}). Each of the variables is represented by a group of related questions, which form a single scale.

\begin{table}[!ht]

\centering
	 \begin{tabular}{| l | l | l | p{1cm} |}
	 \hline
	 Question & Concept & $\mu$ All & $\sigma$ All \\ [0.5ex]
	 \hline\hline
	 Fake/Natural & \textbf{ATPR}  & 3.28 & 0.96	 \\
	 \hline
	 Machinelike/Humanlike &  \textbf{ATPR}  & 2.83 & 0.86	 \\
	 \hline
	 Unconscious/Conscious &  \textbf{ATPR}  & 3.11 & 0.90	 \\
	 \hline
	 Artificial/Lifelike &  \textbf{ATPR}  & 2.94 & 0.80	 \\
	 \hline
	 Dead/Alive & \textbf{ANIM}  & 3.72 & 0.89	 \\
	 \hline
	 Stagnant/Lively & \textbf{ANIM}  & 3.06 & 1.00	 \\
	 \hline
	 Mechanical/Organic & \textbf{ANIM} & 2.56 & 1.04	 \\
	 \hline
	 Artificial/Lifelike & \textbf{ANIM} & 2.78 & 0.81	 \\
	 \hline
	 Inert/Interactive & \textbf{ANIM}  & 3.22 & 1.22	 \\
	 \hline
	 Apathetic/Responsive & \textbf{ANIM}  & 3.61 & 0.70	 \\
	 \hline
	 Dislike/Like & \textbf{LIKB}  & 4.28 & 0.75	 \\
	 \hline
	 Unfriendly/Friendly & \textbf{LIKB}  & 4.50 & 0.71	 \\
	 \hline
	 Unkind/Kind & \textbf{LIKB}  & 4.28 &  0.67	 \\
	 \hline
	 Unpleasant/Pleasant & \textbf{LIKB}  & 4.22 & 0.81	 \\
	 \hline
	 Awful/Nice & \textbf{LIKB}  & 4.39 & 0.70	 \\
	 \hline
	 Incompetent/Competent & \textbf{PINT} & 3.39 & 0.78	 \\
	 \hline
	 Ignorant/Knowledgeable & \textbf{PINT}  & 3.17 & 0.79	 \\
	 \hline
	 Irresponsible/Responsible & \textbf{PINT}  &  3.17 & 0.99	 \\
	 \hline
	 Unintelligent/Intelligent & \textbf{PINT}  & 3.06 & 0.64	 \\
	 \hline
	 Foolish/Sensible & \textbf{PINT}  & 3.50 & 0.51	 \\
	 \hline
	 Anxious/Relaxed& \textbf{PSAF}  & 4.33 & 0.69	 \\
	 \hline
	 Agitated/Calm& \textbf{PSAF}  & 4.22 & 0.88	 \\
	 \hline
	 Quiescent/Surprised& \textbf{PSAF}  & 3.28 & 0.67	 \\
	 \hline
	\end{tabular}
	\caption{\textit{Study B} survey questions and results}
	\label{Tab:tab6}
\end{table}

%% file: A4_sample_conv.tex

\begin{figure}[h]
\centering
\begin{tabular}{p{0.4\linewidth}|p{0.5\linewidth}}
Input & Output\\\hline\hline
\multicolumn{2}{c}{\textit{Example 1}}\\\hline
will it rain today & \cellcolor{wildtalk} i don't like rain \\\hline
\noindent\parbox[c]{\hsize}{\vspace*{1mm}know if it's gonna rain today\vspace*{1mm}} & \cellcolor{wildtalk} yeah it is going to rain \\\hline
\noindent\parbox[c]{\hsize}{\vspace*{1mm}when did was start to rain\vspace*{1mm}} & \cellcolor{genqa} And the answer is 1985\\\hline
\multicolumn{2}{c}{\textit{Example 2}}\\\hline
what is your favorite animal & \cellcolor{genqa} i know for sure, it must be monkey \\\hline
& \cellcolor{persqa} which town do you call your home?\\\hline
hannover & \noindent\parbox[c]{\hsize}{\cellcolor{persqa}\vspace*{1mm} oh, i should visit hannover.\vspace*{1mm}}\\\hline
\multicolumn{2}{c}{\textit{Example 3}}\\\hline
Hello royboy & \cellcolor{hibye!70} greetings \\\hline
how are you & \noindent\parbox[c]{\hsize}{\cellcolor{agentqa} \vspace*{1mm}my lawyer says i don’t have to answer that question.\vspace*{1mm}} \\\hline
\noindent\parbox[c]{\hsize}{\multirow{2}{8em}{What are you up to today}} & \cellcolor{emotion!80} \includegraphics[width=0.18\linewidth]{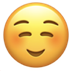} \\\cline{2-2}
& \cellcolor{wildtalk} nothing much \\\hline
& \cellcolor{agentqa} is there anything you want to know about me? \\\hline
How old are you? & \noindent\parbox[c]{\hsize}{\cellcolor{agentqa} \vspace*{1mm}being 1 years old means i am still a child.\vspace*{1mm}}
\end{tabular}
\caption{Our system in a conversation. Color coding: \colorbox{wildtalk}{\module{wildtalk}}, \colorbox{genqa}{\module{genqa}}, \colorbox{persqa}{\module{persqa}}, \colorbox{hibye!70}{\module{hibye}}, \colorbox{agentqa}{\module{agentqa}}, \colorbox{emotion!80}{\module{emotion}}}
\label{fig:study}
\vspace{-10pt}
\end{figure}

Figure \ref{fig:study} shows hand-picked examples from interaction logs of users with \rasta, showcasing specific mechanisms of the \sigslot~automaton as they play out with modules from our reference implementation.\par
While \module{wildtalk} answers the first two questions of \textit{Example~1} using generated responses, \module{genqa} responds to the last one with a general knowledge answer.\par
\textit{Example~2} demonstrates the usage of active engagement. At first, \module{genqa} answers the question from the participant. After a pause in the conversation, proactive engagement (see Appendix \ref{chap:active}) leads to \module{persqa} asking about personal information of the conversation partner. \module{persqa} then stores the answer in a database and generates a positive response to show that the answer was understood.\par
\textit{Example~3} includes multiple modules and active engagement. After \module{hibye} reacts to the greeting, \module{agentqa} answers the question about the dialogue agent. Triggered by the mention of the dialogue agent in the input, \module{emotion} generates a signal for shyness which \module{telegramio} converts into an emoji. In addition to \module{emotion}, \module{wildtalk} reacts to the input and generates a response. \module{agentqa} prompts the user to ask about the dialogue agent as proactive engagement. The corresponding question by the participant is then answered by \module{agentqa}.

%% file: 7_conclusion.tex
\section{Conclusion}

The main contributions of this paper are
\begin{enumerate}
    \item The \sigslot~automaton, a conceptual combination of grammatical symbol systems with slots for intrinsic rule exclusion.
    \item The causal pathway self-information metric which derives an estimate for the prior probability of a signal from the structure of a \sigslot~automaton.
    \item The \rasta software framework for design and deployment of multimodal interactive behavioural policies based on the \sigslot~system, including a structured set of diverse rule modules.
\end{enumerate}
Experimental evidence suggests that the integration of signals, rules and slots adds to the foundation of rule-based interactive systems, improving on its predecessors by simplifying descriptions of concurrent multimodal behaviors. We have shown that \rasta, implementing the \sigslot~system, can effectively employ a contextual selection of task-oriented rules in parallel to specific generative models, enabling effective human-robot interaction policy design. As neural models are reaching human performance on narrow tasks, symbol systems provide an efficient method to implement a switch between them.

\section{Future Work}

In its current form, the \sigslot~system is suitable for modular declarative interaction design. We recognize potential for future work in the system, such as a lack of autonomous long-term goal selection, which prevents the \sigslot~architecture from more direct comparison with mature cognitive architectures such as SOAR. Also, additional experiments could be performed to prove the framework's robustness towards user-initiated context changes. Findings from both user studies also indicate an acute need to improve reliability of human-agent interfaces, such as automatic speech recognition and vision. However, we are confident that further research with the presented system would strengthen the role of hybrid probabilistic symbol systems in interaction policy design, especially when rule formation would be aided by learning \cite{romer2020reinforcement}.

%% file: 8_refs.tex
\printbibliography